\pdfoutput=1

\documentclass[11pt]{article}

\usepackage[dvipsnames, svgnames, x11names]{xcolor}
\usepackage{ulem}

\usepackage[]{emnlp2021}
\usepackage{times}
\usepackage{latexsym}
\usepackage{url}

\usepackage{graphicx}
\graphicspath{ {./images/} }
\usepackage[table,xcdraw]{}
\usepackage{microtype}

\usepackage{booktabs}
\usepackage{multirow} 
\usepackage{makecell}
\usepackage{tabularx}
\usepackage{threeparttable}
\usepackage{amssymb}
\usepackage{marvosym}
\usepackage{subfigure}
\usepackage{soul}

\usepackage[T1]{fontenc}

\usepackage[utf8]{inputenc}

\usepackage{microtype}
\title{Fine-Grained Evaluation for Implicit Discourse  Relation Recognition}

\author{Xinyi Cai\\
 4319cxy
  {\tt @sina.com} \\\
}

\date{}

\begin{document}
\maketitle
\begin{abstract}
Implicit  discourse relation recognition is a challenging task in discourse analysis due to the absence of explicit discourse connectives between spans of text. Recent pre-trained language models have achieved great success on this task. However, there is no fine-grained analysis of the performance of these pre-trained language models for this task. Therefore, the difficulty and possible directions of this task is unclear. In this paper, we deeply analyze the model prediction, attempting to find out the difficulty for the pre-trained language models and the possible directions of this task. In addition to having an in-depth analysis for this task by using pre-trained language models, we semi-manually annotate data to add relatively high-quality data for the relations with few annotated examples in PDTB 3.0. The annotated data significantly help improve implicit discourse relation recognition for level-2 senses.

\end{abstract}


\section{Introduction}
Discourse relations are an important discourse property that embodies the coherence between clauses and sentences. Automatically identifying these discourse relations can help many downstream NLP tasks such as text summarization \cite{cohan-etal-2018-discourse}, text generation \cite{bosselut-etal-2018-discourse}, question answering \cite{jansen-etal-2014-discourse} and machine translation \cite{article-machine-translation}. In the Penn Discourse TreeBank (PDTB) \cite{prasad-etal-2008-penn}, discourse relations are divided into two types: explicit and implicit. While explicit connectives provide strong cues for the models to classify explicit relations, recognizing implicit relations remains a challenging problem due to the  lack of discourse connectives. 

Recent studies on implicit discourse relation recognition have shown success in applying various neural network models including feedforward networks \cite{schenk-etal-2016-really}, convolutional neural networks \cite{zhang-etal-2015-shallow,wang-lan-2016-two}, attention mechanism \cite{liu-li-2016-recognizing,ronnqvist-etal-2017-recurrent}, bidirectional LSTM (Bi-LSTM) \cite{chen-etal-2016-thorough,liu-li-2016-recognizing,dai-huang-2018-improving}, and other neural  methods \cite{qin-etal-2016-shallow,wang-lan-2016-two}. More impressively, pre-trained neural language models have been used and dramatically improved discourse relation recognition \cite{shi-demberg-2019-next,Liu2020OnTI, long2023facilitatingcontrastivelearningdiscourse, long2024leveraginghierarchicalprototypesverbalizer,chan2023discopromptpathpredictionprompt}. 

However, we do not know: which relations the pre-trained language models find hard to recognize; the impact of data distribution on performance of pre-trained language models; the confusing senses pairs for the models; and whether the pre-trained language models would be inconsistent as they separately classify the level-1 and level-2 relations in the same sense hierarchy and so on. 

In this work, we conduct a fine-grained analysis based on the above issues.  
Specifically, our contributions  are: 
\begin{itemize}
\setlength{\itemsep}{0pt}
\setlength{\parsep}{0pt}
\setlength{\parskip}{0pt}
\item  display the difficult and confusing data for the pre-trained language models.

\item reveal the relations between the number of data in PDTB 3.0 \cite{webber2019penn} and the performance of models at level-2 senses, and the types of the level-2 senses whose poor model performance cannot be attributed to the shortage of data.

\item give a linguistic analysis of the model correct and incorrect prediction examples, which demonstrates the pre-trained language models still rely on certain cues to recognize some level-2 senses 

\item  propose a semi-manual annotation approach to augment relatively high-quality data for the relations with few annotated data in PDTB 3.0 \cite{webber2019penn}. And the annotated data  dramatically enhance the F1 score on level-2 discourse relation recognition.

\item finally, we discuss the possible future directions in this task from the perspectives of data and model.

\end{itemize}

\section{Related Work}
There are works evaluating different discourse phenomena. For example, \citet{he-etal-2022-evaluating} evaluate discourse cohesion by proposing a test suite; \citet{wang2023discobenchdiscourseawareevaluationbenchmark} evaluate intra-sentence discourse properties and propose a diagnostic test suite that can examine whether the target models learn discourse knowledge.

The release of the Penn Discourse Treebank (PDTB) \cite{prasad-etal-2008-penn} makes  machine learning based implicit discourse relation recognition more feasible. The task of implicit relation recognition is typically regarded as a classification issue, with the two arguments as input and their implicit discourse relation as the label to predict. Most previous work is based on linguistic and semantic features such as word pairs and Brown cluster pair representation \cite{pitler-etal-2009-automatic,lin-etal-2009-recognizing} or using rule-based systems \cite{wellner-etal-2006-classification}. Recent work has proposed neural network-based models equipped  with attention or advanced representations, such as CNN \cite{qin-etal-2016-shallow}, attention on neural tensor network \cite{guo-etal-2018-implicit}, and memory networks \cite{jia-etal-2018-modeling}. Advanced representations may help to achieve higher performance \cite{bai-zhao-2018-deep}. Some methods also consider paragraph-level context and inter-paragraph dependency \cite{dai-huang-2018-improving}. 

The proposed pre-trained language models lead a great progress for this task. \citet{shi-demberg-2019-next} show that using the bidirectional encoder representation from BERT  \cite{devlin-etal-2019-bert} is effective to capture what events are expected to cause or follow each other. As BERT is trained on a next-sentence prediction task, it encodes a representation of likely next sentences. Their work outperforms previous state-of-the-art approaches in 11-way classification by 8\% points on the standard PDTB dataset. Moreover, \citet{Liu2020OnTI} propose a novel model, BMGF-RoBERTa (Bilateral Matching and Gated Fusion with RoBERTa), which outperforms BERT and other state-of-the-art systems on the standard PDTB dataset around 8\% and CoNLL datasets around 16\%. Recently, \citet{long2023facilitatingcontrastivelearningdiscourse, long2024leveraginghierarchicalprototypesverbalizer} incorporate the sense hierarchy into the recognition process itself and using it to select the negative examples used in contrastive learning, which achieves state-of-the-art performance in this task.

Therefore, conducting a fine-grained analysis on the task of implicit relations recognition by using language models is  necessary and might be helpful for future work.



\section{Dataset}
Penn Discourse TreeBank (PDTB) \cite{prasad-etal-2008-penn} is the largest and most frequently used benchmark dataset for evaluating the recognition of discourse relations. PDTB adopts the predicate-argument structure, where the predicate is the discourse connective (e.g, while) and the arguments are two text spans around the connective. In PDTB, if there is an explicit discourse connective presented in the arguments, the relation between the arguments is explicit; otherwise, it is implicit. In the absence of explicit connectives, identifying the sense of the relations has proved  much more difficult than  the task of explicit relation recognition. 

PDTB has two versions: PDTB 2.0 \cite{prasad-etal-2008-penn} and PDTB 3.0 \cite{webber2019penn}. Version 2.0. of the PDTB, which was released in 2008, contains 40,600 tokens of annotated relations and consists of 2,312 Wall Street Journal articles. It serves as a kind of very useful resources for the development and evaluation of neural models in many downstream NLP applications \cite{narasimhan-barzilay-2015-machine,qin-etal-2017-adversarial,nie-etal-2019-dissent}. And it has also stimulated the development of similar resources in other languages, such as Arabic \cite{AlSaif2010TheLA}, Chinese \cite{Zhou2012PDTBstyleDA}, Czech \cite{Polkov2008FromST}), Italian  \cite{Tonelli2010AnnotationOD}, Turkish \cite{Zeyrek2010TheAS}, and Hindi \cite{Oza2009TheHD}.

The incompleteness and the high value of PDTB 2.0 have led to the appearance of PDTB 3.0, which contains $\sim$13K more tokens annotated for discourse relations, for a total of 53,631 tokens. In addition to more tokens, the sense hierarchy in PDTB 2.0 has been revised and much more intra-sentential relations have been annotated  in PDTB 3.0. 
 With the release of PDTB 3.0, there has been several annotation work in PDTB style \cite{Zeyrek2020TEDMD,Long2020TEDCDBAL,long2020shallowdiscourseannotationchinese} following the settings of PDTB 3.0.  

Considering the size and the improvement of PDTB.3.0, our work adopts it as the evaluation dataset. All PDTB relations in PDTB 3.0 are hierarchically categorized into 4 level-1 classes: Expansion, Comparison, Contingency, and Temporal, which are further divided into 17 level-2 types. To have a better understanding of the model and data for the task of implicit relation recognition, we evaluate models on the level-2 implicit types.



\begin{table*}
\setlength{\belowcaptionskip}{-0.5cm}
\small
\centering
\begin{tabular}{|l|l|l|l|l|}
\bottomrule
 Class&Size&BERT (large)&
ALBERT (xlarge)&RoBERTa (large)\\
\hline
Temp.Synchronous&539&27.01 ($\pm$6.26)&12.05 ($\pm$7.69)&47.13 ($\pm$5.72)\\
Temp.Asynchronous&1,286&55.11 ($\pm$3.30)&49.74 ($\pm$7.93)&65.83 ($\pm$3.96)\\
Cont.Cause&5,785&{\color{RoyalBlue}\textbf{62.51 ($\pm$1.31))}}&{\color{RoyalBlue}\textbf{60.75 ($\pm$3.03)}}&70.33 ($\pm$1.38)\\
Cont.Cause+Belief&202&16.05 ($\pm$9.88)&8.15 ($\pm$9.77)&\color{BrickRed}{\textbf{15.84 ($\pm$11.62)}}\\
{\color{BurntOrange}\textbf{Cont.Condition}}&198&61.44 ($\pm$9.47)&{\color{BrickRed}\textbf{2.90 ($\pm$6.48)}}&{\color{RoyalBlue}\textbf{72.17 ($\pm$7.38)}}\\
Cont.Purpose&1,352&{\color{RoyalBlue}\textbf{90.87 ($\pm$2.25)}}&{\color{RoyalBlue}\textbf{87.20 ($\pm$2.20)}}&{\color{RoyalBlue}\textbf{92.40 ($\pm$1.22)}}\\
Comp.Concession&1,494&40.63 ($\pm$3.35)&30.28 ($\pm$13.29)&59.81 ($\pm$2.63)\\
Comp.Contrast&983&41.89 ($\pm$3.70)&33.37 ($\pm$14.83)&57.97 ($\pm$3.56))\\
Comp.Similarity&31&{\color{BrickRed}\textbf{10.00 ($\pm$21.08)}}&	10.00 ($\pm$21.08)&{\color{BrickRed}\textbf{10.00 ($\pm$21.08)}}\\
Exp.Conjunction&4,376&57.16 ($\pm$1.54)&58.19 ($\pm$2.82)&66.07 ($\pm$1.35)\\
Exp.Disjunction&30&{\color{BrickRed}\textbf{0.00 ($\pm$0.00)}}&{\color{BrickRed}\textbf{0.00 ($\pm$0.00)}}&{\color{BrickRed}\textbf{0.00 ($\pm$0.00)}}\\
Exp.Equivalence&336&{\color{BrickRed}\textbf{0.98 ($\pm$3.09)}}&{\color{BrickRed}\textbf{0.98 ($\pm$3.09)}}&19.10 ($\pm$9.77)\\
Exp.Instantiation&1,533&53.85 ($\pm$1.82)&49.90 ($\pm$5.09)&62.85 ($\pm$1.80)\\
{\color{BurntOrange}\textbf{Exp.Level-of-detail}}&3,361&47.60 ($\pm$1.63)&43.31 ($\pm$3.79)&57.74 ($\pm$1.99)\\
Exp.Manner&726&{\color{RoyalBlue}\textbf{78.22 ($\pm$5.23)}}&{\color{RoyalBlue}\textbf{69.63 ($\pm$3.72)}}&{\color{RoyalBlue}\textbf{80.52 ($\pm$5.02)}}\\
Exp.Substitution&450&57.76 ($\pm$4.23)&20.81 ($\pm$12.90)&71.40 ($\pm$6.33)\\
\hline
Macro F1&-&43.82($\pm$1.57)&33.58($\pm$3.83)&53.07($\pm$1.73)\\
\toprule
\end{tabular}
\caption
{F1 score of each level-2 relation and total macro F1 score and corresponding standard deviation for the level-2 relation recognition on the PDTB 3.0 dataset.}
\label{table-2}
\end{table*}
\section{Experiment Setup}
We used  three strong sentence encoder models  as our baselines: BERT \cite{devlin-etal-2019-bert}, ALBERT \cite{lan2019albert} and RoBERTa \cite{liu2019roberta} which have been used in many NLP tasks \cite{he2025evaluatingimprovinggraphtext,long-etal-2020-ted, long-etal-2024-multi,he2025mintqamultihopquestionanswering}. And we use a publicly available codebase\footnote{https://github.com/huggingface/transformers}. To fine-tune on PDTB 3.0, we use the final hidden vector
$C\epsilon R^H$ corresponding to the first input token ([CLS]) as the aggregate representation. The only new parameters introduced during fine-tuning are classification layer weights $W\epsilon R^{K X H}$, where K is the number of labels. We compute a standard classification loss with C and W. For all sentence encoder models, we fine-tuned each model for a maximum of 10 epochs with early stopping when the the development set performance did not improve for 5 evaluation steps with a batch size of 8. We used a learning rate of 2e-6 for all models. Accuracy were used as the validation metrics. Following \citet{shi-demberg-2017-need}, we treated instances with multiple annotations as separate examples during training. A prediction is considered correct if it matches any of the labels during testing. For classification, we used the standard split treating training vs. dev vs. test as 8:1:1. We adopted 10-fold cross-validation for the level-1 and level-2 classification, sharing the concerns of \citet{shi-demberg-2017-need} on label sparsity. We used cross-validation at the individual example level for comparing with the results with extra data. Our baselines showed strong performance on all splits.

\begin{table*}
\setlength{\belowcaptionskip}{-0.5cm}
\small
\centering
\begin{tabular}{|l|l|l|l|l|l|l|l|}
\bottomrule 
\multirow{2}{*}{\qquad\quad Class}&
\multirow{2}{*}{\qquad Size}&\multicolumn{2}{c|}{BERT (large)}&
\multicolumn{2}{c|}{ALBERT (xlarge)}&\multicolumn{2}{c|}{RoBERTa (large)}\\
\cline{3-4}\cline{5-6}\cline{7-8}
&&\qquad\quad -&\quad$\Delta$&\qquad\quad-&\quad$\Delta$&\qquad\quad-&\quad$\Delta$\\
\hline
Synchronous&2,039 (+1,500)&55.72 ($\pm$2.11)&+28.71&20.84 ($\pm$13.3)&+8.79&64.85 ($\pm$1.53)&+17.72\\
Asynchronous&2,786 (+1,500)&60.31 ($\pm$2.79)&+5.29&48.37 ($\pm$7.63)& -1.37&70.21 ($\pm$2.68)&+4.38\\
Cause&5,785 (+0)&	58.73 ($\pm$1.46)&-3.78&	52.56 ($\pm$5.67)&-8.19&67.91 ($\pm$1.52)&-2.42\\
Cause+Belief&414 (+212)&38.48 ($\pm$8.59)& +22.43&42.85 ($\pm$15.46)&+34.70&53.79 ($\pm$9.46)&+37.95\\
Condition&198 (+0)&	59.96 ($\pm$11.7)&-1.48&0.00 ($\pm$0.00)&-2.90&72.87 ($\pm$8.13)&+0.70\\
Purpose&1,487 (+135)&89.42 ($\pm$1.60)&-1.45&80.34 ($\pm$9.49))&-6.86&91.04 ($\pm$2.03)&-1.36\\
Concession&2,994 (+1,500)&51.04 ($\pm$1.66)&+10.41&39.24 ($\pm$14.95)&8.96&64.96 ($\pm$2.28)&+5.15\\
Contrast&2,148 (+1,165)&52.82 ($\pm$1.28)&+10.93&32.89 ($\pm$20.14)&-0.48&65.55 ($\pm$2.55)&+7.58\\
Similarity&130 (+99)&2.67 ($\pm$8.43)&{\color{BrickRed}\textbf{-7.33}}&	2.67 ($\pm$8.43)&{\color{BrickRed}\textbf{-7.33}}&11.60 ($\pm$16.27)&+1.60\\
Conjunction&4,376 (+0)&53.12 ($\pm$2.58)&-4.04&50.52 ($\pm$6.25)&-7.67&64.37 ($\pm$1.44)&-1.70\\
Disjunction&251 (+221)&11.11 ($\pm$8.09)&+11.11&6.53 ($\pm$7.93)&+6.53&48.28 ($\pm$8.00)&+48.28\\
Equivalence&562 (+226)&34.84 ($\pm$4.18)&+33.86&30.35 ($\pm$13.55)&+29.37&43.16 ($\pm$5.75)&+24.06\\
Instantiation&1,820 (+287)&54.56 ($\pm$3.26)&+0.71&49.11 ($\pm$8.49)&-0.79&67.18 ($\pm$2.49)&+4.33\\
Level-of-detail&3,361 (+0)&45.65 ($\pm$2.16)&-1.95&36.59 ($\pm$7.85)&-6.72&57.26 ($\pm$1.56)&-0.48\\
Manner&726 (+0)&78.11 ($\pm$2.91)&-0.11&68.17 ($\pm$8.19)& -1.46&80.60 ($\pm$5.31)&+0.08\\
Substitution&707 (+257)&53.12 ($\pm$7.94)&{\color{BrickRed}\textbf{-4.64}}&18.22 ($\pm$9.93)&{\color{BrickRed}\textbf{-2.59}}&68.36 ($\pm$3.8)&{\color{BrickRed}\textbf{-3.04}}\\
\hline
Macro F1&-&49.98 ($\pm$1.45)& +6.16&36.20 ($\pm$7.76)&+2.62&62.0 ($\pm$1.02)&+8.93\\
\toprule
\end{tabular}
\caption
{F1 score of each level-2 relation and total macro F1 score and corresponding standard deviation for the level-2 relation recogonition  on the combination of  PDTB 3.0. The bracketed number for each level-2 relation denotes the number of instances newly added.}
\label{table-3}
\end{table*}

\section{Experiments and Analysis on level-2 Senses}
In this section, we present the experimental results of implicit relation recognition for  level-2 senses by using pre-trained language models. It is known that the PDTB-3 sense hierarchy has 4 top-level senses (Expansion, Temporal, Contingency, Contrast) and 17 second-level senses. To explore the difficulty improve the performance of this task, it is necessary to investigate how models perform for the second-level senses. Taking into account the impact of data distribution and the similarities between some senses, we try to reveal 1)To improve the performance of the models, which level-2 senses are subject to the insufficient data; 2) what level-2 senses are difficult for the models, when data is not a primary problem; 3) which pairs of level-2 senses are confusing for the pre-trained language models.
\subsection{Data Distribution and Experimental Results}

It cannot be denied that the lack of data for some senses must lead to the poor performance for them and that the imbalanced data distribution is a common issue for most corpora. However, the poor performance for all senses can not only be attributed to this same reason, which might arise in the difficulty of the senses themselves.

Therefore, in this section, we investigate the correlation between the number of data for each type of level-2 senses and the model performance for them, trying to find out the
find out the difficult level-2 senses for the models and exploring whether the difficulties for different kinds of senses arise out of insufficient data.


Table \ref{table-2} show the experimental results of three pre-trained language models for level-2 relation classification. It should be noted that here we exclude Exception, for which has less than\textless 30 labelled instances, so the table displays 16 level-2 senses. 

Firstly, we can conclude that Purpose, Manner, and Condition are not difficult for these pre-trained language models, as their F1 scores are 90.87\%, 80.52\%, and 72.17\%, which are much better than that of other senses, while their data sizes are not large, with 1,352, 450, and 198 respectively. By contrast, although there are 3,341 labelled examples for the sense of Level-of-detail, the pre-trained language models only get the F1-score around 45\% to 55\%. Therefore, Level-of-detail is quite challenging for the models. Besides, the poor performance of the models for Similarity, Cause.belief, Disjunction might be attributed to the shortage of the data, for which only have 31, 202 and 30 instances respectively among the dataset of PDTB 3.0. Besides,  the number of annotated instances for Cause is the largest, with 5785, but the models do not perform best for it, so to some degree indicate its difficulty for the models.


\subsection{New data and Experimental Results}
To learn whether the model can perform better for level-2 senses if more data are provided, we semi-manually annotated data to augment relatively high-quality data for the relations with \textless 1,000 instances in PDTB 3.0. As good performance has been achieved for Condition, we did not have additional annotated data for this relation. The method we adopted is semi-manual annotation. It was not too expensive, but the annotation quality can be guaranteed.
 
The source data for our augmentation is from the explicit relations in both PDTB 3.0 and Kaggle TED Talks Dataset\footnote{ https://www.kaggle.com/rounakbanik/ted-talks}. With the intention to balance the dataset, we control the number of added data for each category by not only ensuring that the added data for each would not exceed 1,500 but also rendering the size of the relations with few instances as large as possible. 

As for the annotation methods, it is motivated by the methods adopted to automatically extend PDTB \cite{Braud2014CombiningNA,Marcu2002AnUA}. We designed a pipeline that automatically  extracts discourse relation argument pairs through taking advantage of the connective words that are known as clear relation indicators and then manually annotates level-2 discourse labels for the extracted argument pairs. In the first step, all instances with explicit discourse relations are automatically extracted. In the second step, what we need to do is removing the connectives. However, because the original meaning would be changed \cite{sporleder_lascarides_2008} if some connectives are removed, we follow the method proposed by \citet{rutherford-xue-2015-improving} to identify the connectives which are freely omissible by measuring the Omissible Rate and Context Differential. The selected connectives include similarly, for instance, for example, in other words, etc. 
\begin{figure*}[!tbp]
\flushleft
\includegraphics[scale=0.185]{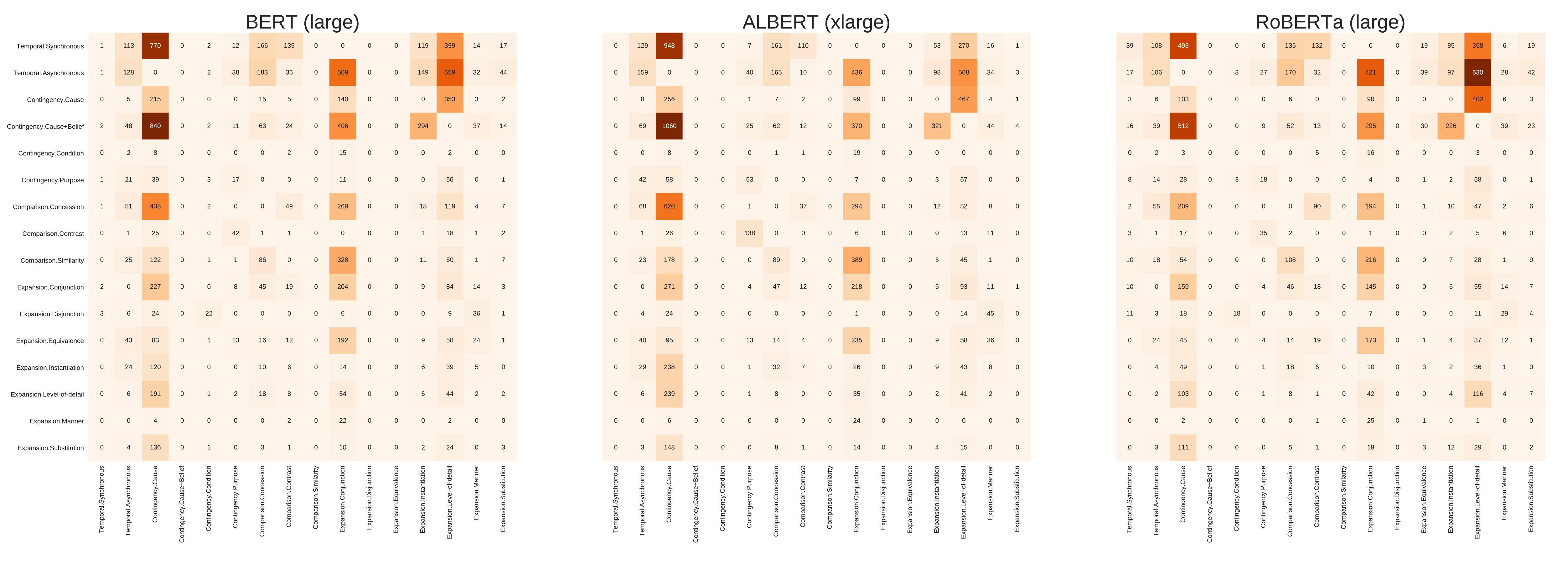}
\caption
{Confusion matrice for level-2 senses on PDTB 3.0.}
\label{figure-1}
\end{figure*}

Then, unlike previous work, as we target the level-2 discourse relations instead of the four level-1 relations, it is impossible that discourse relation labels at this level can be automatically assigned so far. Hence, manual work is required after the argument pairs are extracted. We hire two experienced annotators to annotate these extracted pairs. They are able to annotate the relations more quickly with the assistance of the “removed connectives”. To ensure the manual annotation quality, before annotation, the annotators studied the annotation guidelines and discussed together. Then, 300 examples were used as samples for the agreement study, and the agreement rate between the annotators are 0.80.  Finally, the total number of annotated relations is 7,102, which took the two annotators over one month to annotate and check.

We conducted experiments on the data from PDTB 3.0 plus our additionally annotated data using the same models with the same parameters. The results are displayed in Table \ref{table-3}. Obviously, compared with the results shown in Table \ref{table-2}, the overall F1 score are significantly improved with the help of newly added data, which fully demonstrate our new data is helpful and our semi-manually annotated method is useful.

On the one hand, we can see that for some senses like Disjunction, Cause. belief and Equivalence, the model can perform much better for them, when more data are provided, as their F1 score of RoBERTa dramatically are increased by 48.28\%, 37.95\% and 24.06\% respectively. Therefore, while manual annotation is very expensive and time-consuming, we can consider to annotate more data for the level-2 senses whose model.
performance are largely affected by a lack of data.

On the other hand, with more data added, the performance of the models do not become better or even worse for substitution and similarity . To know the reason, we double- checked our annotation examples of these two senses, and found there is no problems in our annotation. Therefore, these two senses might be difficult due to the features of themselves rather than being affected by not enough data. Both substitution and similarity are under the category of Expansion, so helping the models to identify the features of these two level-2 senses would also benefit the model in the top level classification(4-way classification).

\subsection{Confusing Level-2 Senses}
Figure \ref{figure-1} shows the confusion matrices for predictions on level-2 senses of PDTB 3.0, from BERT-large, ALBERT-xlarge and RoBERTa-large. The figures aggregate the predictions from all test sets of the cross-validation experiment. Colors in figures are normalized over each row, telling us the frequency when a relation denoted in the row is misclassified into another wrong relation denoted in the column. Looking at the part with deeper color in these figures, we can know that these pairs of level-2 senses are confusing for all of these three models, which are Cause.belief\&Cause, Synchronous\/Cause, Asynchronous\&Conjunction, Conjunction\&Synchronous, Synchronous\&Level of detail, asynchronous\&Level of Detail, Cause\&Level of Detail, Instantiation\&Cause.belief.

Cause+belief and Cause is quite similar, so it can be expected that they are confusing for the models, and the reason why the model cannot distinguish them is that the model cannot identify whether one argument is a claim or not. 
\begin{table*}
\scriptsize
\centering
\begin{tabular}{ccp{22em}ccc}
\toprule
Cues&location&Examples&Level-2 Senses&Correct&Wrong\\
\hline
\multirow{4}{*}{To-infinitives}&\multirow{4}{*}{Arg2}&[Mr. Jackson showed up at the affair as well]$_{Arg1}$ [\underline{to sign autographs} for a fee.]$_{Arg2}$&Purpose	&98.70\%&46.50\%\\
		&&[Now, they'll have to increase their coffers]$_{Arg1}$ [\underline{to protect for the future}.]$_{Arg2}$	&Condition	&98.90\%&	70.00\%\\
\hline
\multirow{5}{*}{And+verb}&	Arg2	&[toss him out of the building.]$_{Arg1}$ [\underline{and let} the forces of the status quo explain to the parents whatever it is they're defending.]$_{Arg2}$ &	Asychronous	&62.40\%	&4.70\%\\
	&Arg2	&[throw away the passbook]$_{Arg1}$ [\underline{and go} for the glory.]$_{Arg2}$ 	&Substitution& 	4.50\%&	17.00\%\\
\hline
\multirow{4}{*}{Negation}	&Arg1	&[Japan is \underline{not} a political country.]$_{Arg1}$ [It is a bureaucratic country.]$_{Arg2}$ &	Substitution &	93.90\%	&49.60\%\\
	&Arg2&	[In the past, customers had to go to IBM when they outgrew the VAX.]$_{Arg1}$ [now they \underline{don't} have to.]$_{Arg2}$ &	Concession&	84.20\%	&18.50\%\\
\hline
Same subject &	Arg1 and Arg2&	[\underline{it's} no longer news.]$_{Arg1}$ [\underline{it's} drama.]$_{Arg2}$ &	Substitution 	&61.30\%	&5.60\%\\
\hline
\multirow{2}{*}{Similar syntactic pattern} &	\multirow{2}{*}{Arg1 and Arg2}&	[\underline{Apple Computer gained 1 to 48 1/4}.]$_{Arg1}$ [\underline{Ashton-Tate rose 3/8 to 10 3/8}.]$_{Arg2}$ &	\multirow{2}{*}{Conjunction}&\multirow{2}{*}{53.50\%}&\multirow{2}{*}{4.30\%}\\

\bottomrule
\end{tabular}
\caption
{Linguistic cues and their percentage in correct and wrong model prediction at some level-2 senses.}
\label{table-linguistic}
\end{table*}

Moreover, when the relations are Temporal, which can be Synchronous or Asynchronous, models often cannot recognize and tend to identify  
Synchronous as Cause and Asynchronous/conjunction. Therefore, the models have difficulty in distinguish temporal relations from Cause and Conjunction.

Besides, many relations are more likely to be wrongly classified into Conjunction, Level of Detail and Cause, for their number of annotated examples account for 19.29\%, 14.81\%, and 25.50\% respectively. The large percentage of these senses negatively the performance of the models, because when it is difficult for the models to decide the sense of an example, they are prone to assign the labels of these senses to it.

\section{Linguistic Analysis of Model Prediction}

We present a fine-grained linguistic analysis of model prediction in this section. When trying to analyze the correct and incorrect prediction examples, we noticed that there are strong surface linguistic cues for the examples of some level-2 senses. Some surface cues can only be found for most of level-2 senses in PDTB 3.0, which are likely not too conspicuous in PDTB 2.0 since  $\sim$ 13k additional intra-sentential relations have been annotated in the new version of PDTB, including free adjuncts, free to-infinitives prepositional clausal subordinates and conjoined verb phrases.

After observing the prediction results and computing their frequency in some level-2 senses, we select five linguistic cues to display, and they are to-infinitive, and+verb, negation, same subject and similar syntactic pattern. Table 4 provides examples for each of these cues. What we want to do is to investigate whether and how much the model depend on these cues to recognize certain types of level-2 senses, and we want to know whether the model can still perform well,  without these cues. 




\noindent
\textbf{To-infinitives} Free to-infinitives have been annotated as implicit relations in PDTB 3.0. While the sense ascribed to these relations is most often Contingency. Purpose, it can be seen from Table \ref{table-linguistic} that almost all correctly predicted examples for Purpose and Condition have to-infinitives, with 98.7\% and 98.9\% respectively, while this linguistic cue appear in much less wrong prediction examples. This illustrate that the good performance of models on these two senses greatly depend on this surface cues.\\
\noindent
\textbf{And+verb} And+verb, is usually parsed as a kind of conjoined VPs in the Penn TreeBank. The  argument that begins with this structure is labeled as Arg2. This cue can be mainly found in the relation of Temporal. Asynchronous. Comparing the percentages in the correct and wrong predictions, we can find that the model is good at utilizing this surface feature and tend to associate it as Temporal.Asynchronous, because we can see this cue can be seen in 62.7\% right prediction of Asynchronous and 4.7\% of wrong prediction.\\
\noindent
\textbf{Negation} Negation is usually used to express the contradiction towards the same topic. From the percentages shown in the Table \ref{table-linguistic}, the percentage of the correct prediction examples having negation is 93.90 \% for Substitution and 84.20 \% for Concession, so the model rely excessively on negation when recognizing these two senses. However, if there is no negation, it is much difficult for the model to identify these senses, for most of the wrong prediction do not have negation.\\
\noindent
\textbf{Same subject} If the two subjects in the arguments are coreferential, the two subjects are said to be the same. We notice that Substitution relations usually have the same subjects in the correct predictions, accounting for 61.3\%, while there are few examples in the wrong predictions having such features, with 5.6\%, which to some extent reveals that the model encodes knowledge about coreference well and often associate it with Substitution.
\begin{figure*}[!tbp]
\flushleft
\includegraphics[scale=0.18]{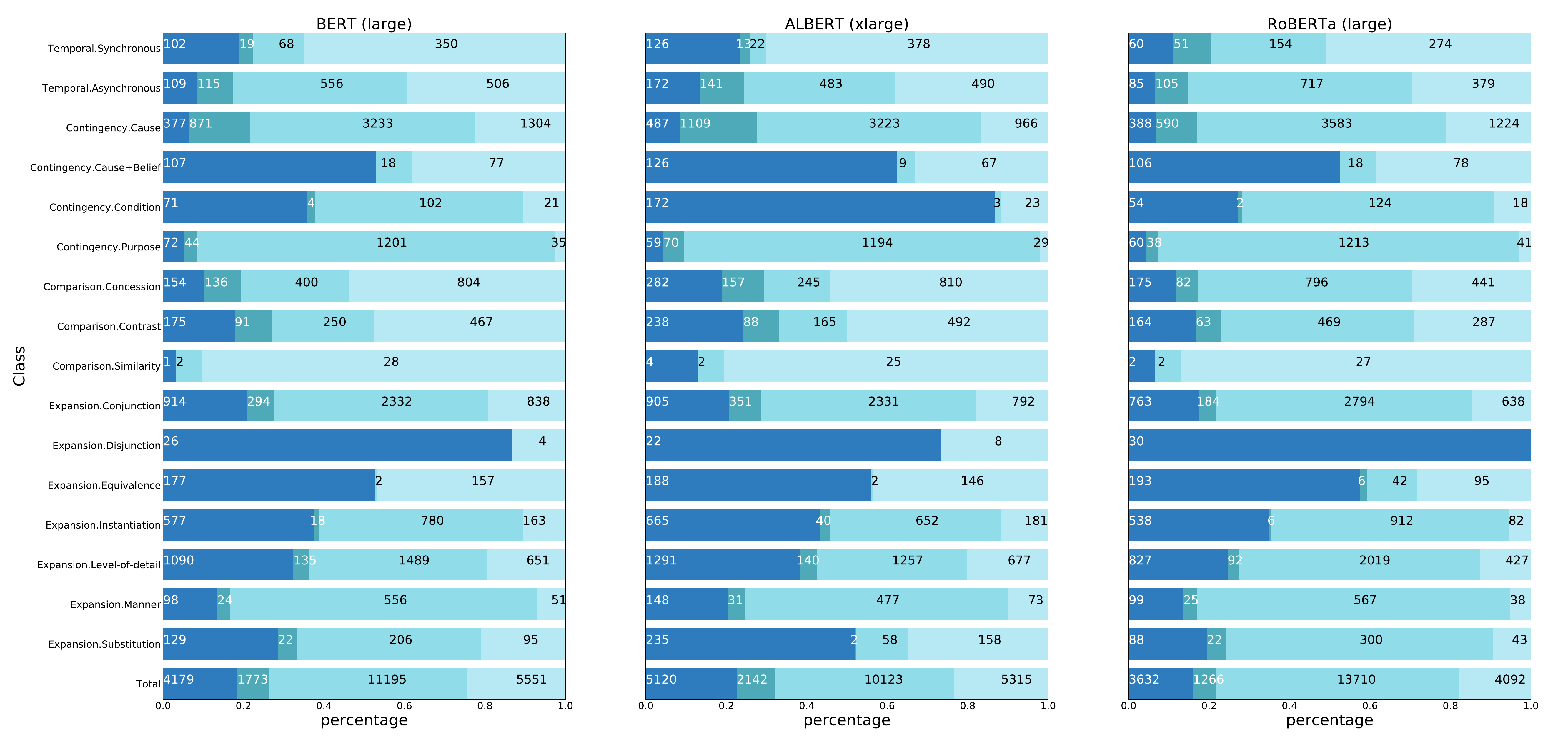}
\caption
{Inconsistent model predictions across sense levels. The color from left to right indicate that the predictions are: correct for level-1 senses but wrong for level-2 senses; correct for level-2 senses but wrong for level-1 senses; correct for both level-1 and level-2 senses; wrong for both level-1 and level-2 senses.}
\label{figure-2}
\end{figure*}

\noindent
\textbf{Similar syntactic pattern}: \citet{lin-etal-2009-recognizing} argue that syntactic structure within one argument may constrain the relation type and the syntactic structure on the other argument. From our observation of the model prediction for Expansion.Conjunction, most correct model predictions have similar syntactic  structures. We can see that both arguments in the example  shown in Table \ref{table-linguistic} is the structure of NP+VP+NP. We  manually evaluated 400 correct prediction examples and 400 wrong ones for Conjunction, finding that 53.5\% of correct ones have similar syntactic patterns in the two arguments and almost 96\% wrong prediction examples do not share similar syntactic patterns between arguments. This suggests that the model utilize this surface syntactic knowledge for predicting Conjunction.

In short, this section displays that even the pre-trained language models dramatically increase the performance of the task of implicit relation recognition, we stii find that for some level-2 senses, if there are no certain linguistic features in the examples, the models would perform much more badly.


\section{Inconsistent Model Prediction Across Sense Levels}
Previous work on implicit relation recognition generally carry out sense classifications at each level separately, not taking the mapping relationships between adjacent-level senses into account. However, the PDTB 3.0 sense hierarchy indicates there are four main level-1 discourse senses and each of them is further refined into several level-2 discourse senses. In other words, there is a one-to-many mapping between senses at the level-1 and those at the level-2. Therefore, we want to study whether the predictions across different levels are consistent.

As shown in the following example, when the models classify the level-1 relation correctly, but misclassify the corresponding level-2 relation to other relations that are not under the predicted level-1  relation, or when the models classify the level-2 relation correctly but not for the level-1 relation, we see it as inconsistent prediction across sense levels. 
\begin{itemize}
\setlength{\itemsep}{0pt}
\setlength{\parsep}{0pt}
\setlength{\parskip}{0pt}
\item[] [Many people in Poland hope this government will break down.]$_{Arg1}$ [That's what the naczelnik counts on.]$_{Arg2}$\\
Correct: Expansion $\rightarrow$ Conjunction. \\
Inconsistent: Expansion $\rightarrow$ Cause \\
Inconsistent: Contingency $\rightarrow$ Conjunction 
\end{itemize}

We firstly conduced experiments of the level-1 implicit relation recognition, and we got the correct and incorrect prediction examples. Then the experiments of level-2 implicit relation recognition were done and also got the correct and incorrect prediction examples on this level. Finally, we calculate the number of the  examples in terms of the following four  :  correct for level-1 senses but wrong for level-2 senses; correct for level-2 senses but wrong for level-1senses; correct for both level-1 and level-2 senses; wrong for both level-1 and level-2 senses. And we also use BERT-large, AlBERT-large, and RoBerta-large for these experiments.

Figure \ref{figure-2} illustrates the inconsistent model predictions between the two-level senses. Most are cases that models correctly predict level-one senses but wrongly for level-2 senses. In addition,  we can also see that sometimes even the models correctly identify Contingency.Cause and Temporal.Asynchronous for 16-way level-2 classification, they are also likely to fail in the 4-way level-1 classification.

Therefore, although separately evaluating models on level-1 and level-2 implicit relations recognition seems to be more convenient, but the inconsistent prediction of the same models across levels indicate the necessity to more fully use the hierarchical information between level-1 and level-2.

\section{Future: Data and Models}
In this section, we conclude what we can do next from the perspectives of data and model.

\noindent
\textbf{Data}
\begin{itemize}
\setlength{\itemsep}{0pt}
\setlength{\parsep}{0pt}
\setlength{\parskip}{0pt}
    \item Data augmentation for only some level-2 senses. To further improve the performance of models on this task, more data is needed. However, the difficulty for some relations have nothing to do with the number of annotated examples (see section 5.1 and 5.2) and too much certain types of data also may mislead the model (see section 5.3), it would probably better for us only to augment the data for some Level-2 senses with few annotated labels, such as Disjunction, Equivalence, and Cause+Belief, etc. 
    \item Better annotation methods. While Manual data annotation is costly and time-consuming and the quality of fully automatic annotation cannot be guaranteed, we think the best way is to make human annotation less time-consuming and convenient by fully taking advantage of the automatic methods instead of only using human annotation or automatic annotation. Our proposed semi-manually annotation is an example, but better similar annotation methods can be explored.
    \item Extract examples from more texts of different genres. PDTB and all previous PDTB-style dataset annotated data for a small number of texts from the same genre like news or Ted talks, but this gives rise to the imbalanced data distribution and limited diversity of the data. Therefore, to enable the models to learn more, based on the features of different senses, we can extract examples from more texts, and the texts can be spoken or written, and they can be from different genres. 
 
\end{itemize}
\textbf{Model}
\begin{itemize}
\setlength{\itemsep}{0pt}
\setlength{\parsep}{0pt}
\setlength{\parskip}{0pt}
    \item Focus on difficult level-2 senses. Through our analysis, we know what data are truly difficult and confusing for the models. To further improve the models, what we can do is to further explore how to make the models better understand the difficult data. For example, in section 5.3, we mentioned that the models tend to identify Synchronous as Cause and Asynchronous as conjunction, which indicate the models need to capture more temporal knowledge. 
    \item Test on difficult data. Our linguistic analysis of the model prediction show that the great model performance for some types of level-2 senses excessively reply on certain surface linguistic cues, so it would be good to exclude those examples containing the surface linguistic cues and guarantee the diversity of the test set. Only in this way, can we better evaluate the models and know where to improve more clearly. 
    \item Utilize the hierarchical information between levels. Instead of separately evaluating the models on level-1 and level-2 implicit relation recognition, our analysis about inconsistent prediction across levels encourage future models to consider using the hierarchical information by proposing models that can predict the level-1 senses and level-2 senses simultaneously.

\end{itemize}

\section{Conclusion}
In this paper, we have conducted a fine-grained analysis for the task of implicit discourse relation recognition by using the state-of-the-art pre-trained language models. Our in-depth annalysis reveal the prossble directions from the pespectives of data and models including including data augmentation for only some level-2 senses, extracting examples from more texts of different genres, focusing on difficult level-2 senses, testing  on difficult data, and utilizing hierarchical information between level-1 and level-2.

\normalem
\bibliography{anthology,custom}
\bibliographystyle{acl_natbib}




\end{document}